\documentclass[sigconf]{acmart}

\usepackage{subfigure}

\usepackage{amssymb}
\usepackage{multirow}

\AtBeginDocument{%
  \providecommand\BibTeX{{%
    \normalfont B\kern-0.5em{\scshape i\kern-0.25em b}\kern-0.8em\TeX}}}

\copyrightyear{2020}
\acmYear{2020}
\setcopyright{acmcopyright}
\acmConference[MM '20]{Proceedings of the 28th ACM International Conference on Multimedia}{October 12--16, 2020}{Seattle, WA, USA}
\acmBooktitle{Proceedings of the 28th ACM International Conference on Multimedia (MM '20), October 12--16, 2020, Seattle, WA, USA}
\acmPrice{15.00}
\acmDOI{10.1145/3394171.3413666}
\acmISBN{978-1-4503-7988-5/20/10}

\begin{document}
\fancyhead{}

\title{Spatio-Temporal Inception Graph Convolutional Networks \\ for Skeleton-Based Action Recognition}


\author{Zhen Huang}
\authornote{This work was done when the author was visiting Alibaba as a
research intern.}
\affiliation{\institution{University of Science and Technology of China}}
\email{hz13@mail.ustc.edu.cn}

\author{Xu Shen}
\affiliation{\institution{Alibaba Group}}
\email{shenxu.sx@alibaba-inc.com}

\author{Xinmei Tian}
\authornote{Corresponding authors.}
\affiliation{\institution{University of Science and Technology of China}}
\email{xinmei@ustc.edu.cn}

\author{Houqiang Li}
\affiliation{\institution{University of Science and Technology of China}}
\email{lihq@ustc.edu.cn}

\author{Jianqiang Huang}
\affiliation{\institution{Alibaba Group}}
\email{jianqiang.hjq@alibaba-inc.com}

\author{Xian-Sheng Hua}
\authornotemark[2]
\affiliation{\institution{Alibaba Group}}
\email{xiansheng.hxs@alibaba-inc.com}


\begin{abstract}
Skeleton-based human action recognition has attracted much attention with the prevalence of accessible depth sensors. Recently, graph convolutional
networks (GCNs) have been widely used for this task due to their powerful capability to model graph data. The topology of the adjacency graph is a key factor for modeling the correlations of the input skeletons. Thus, previous methods mainly focus on the design/learning of the graph topology. But once the topology is learned, only a single-scale feature and one transformation exist in each layer of the networks. Many insights, such as multi-scale information and multiple sets of transformations, that have been proven to be very effective in convolutional neural networks (CNNs), have not been investigated in GCNs. The reason is that, due to the gap between graph-structured skeleton data and conventional image/video data, it is very challenging to embed these insights into GCNs. To overcome this gap, we reinvent the split-transform-merge strategy in GCNs for skeleton sequence processing. Specifically, we design a simple and highly modularized graph convolutional network architecture for skeleton-based action recognition.
Our network is constructed by repeating a building block that aggregates multi-granularity information from both the spatial and temporal paths. Extensive experiments demonstrate that our network outperforms state-of-the-art methods by a significant margin with only $1/5$ of the parameters and $1/10$ of the FLOPs.
Code is available at \url{https://github.com/yellowtownhz/STIGCN}.
\end{abstract}

\maketitle

\section{Introduction}
\label{sec:introduction}

\begin{figure}[!htbp]
  \centering
  \subfigure[GoogLeNet Inception for CNNs]
  {
      \includegraphics[width=0.7\linewidth]{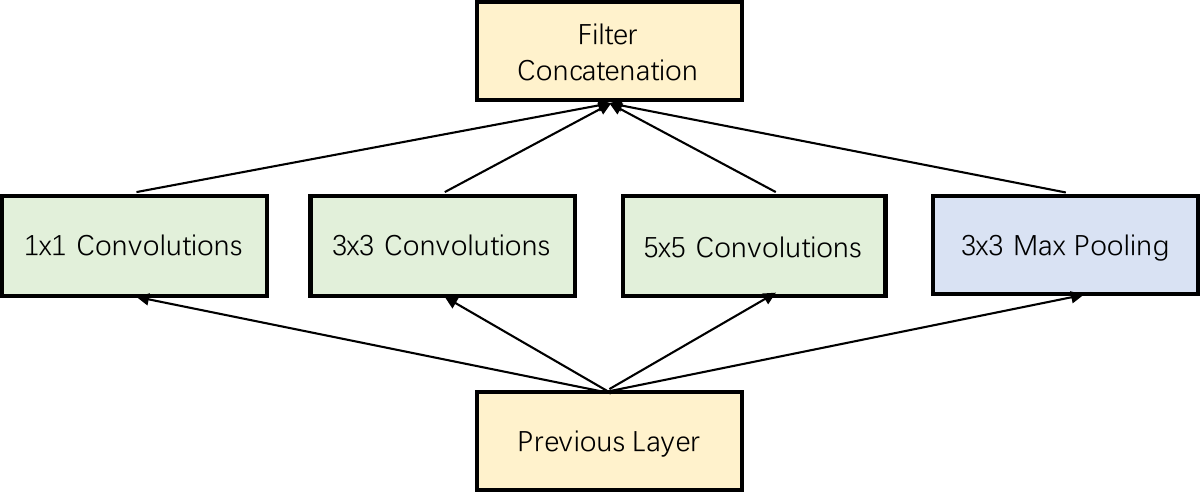}
  }
  \subfigure[Our Spatio-Temporal Inception for GCNs]
  {
    \includegraphics[width=\linewidth]{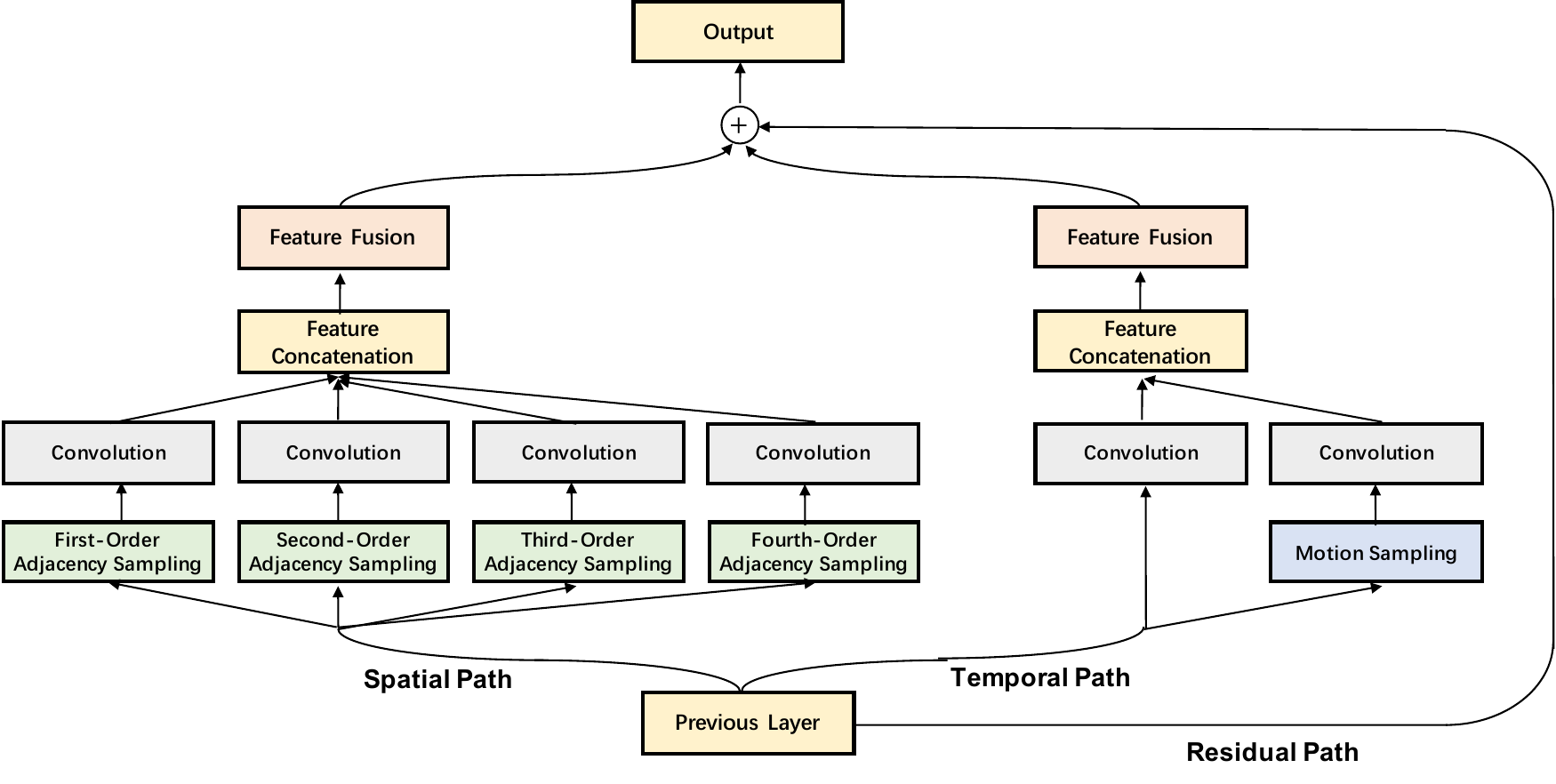}
  }
  \vspace{-5mm}
  \caption{GoogLeNet inception block for CNNs and our spatio-temporal inception block for GCNs.
  Both blocks follow the split-transform-merge strategy.
  In our spatio-temporal (ST) inception block, the inputs are split into three
  paths: a spatial path for spatial features, a temporal path for sequential
  features, and a residual path for the reuse of the input features. In the spatial
  path, graph convolutions with $1\times$ to $4\times$ hop connections are
  applied. In the temporal path, graph convolutions are applied to
  position features and motion features.}
  \label{fig:fusion}
  \vspace{-5.5mm}
\end{figure}

Human action recognition attracts considerable attention due to their
potential for many applications.
Recently, skeleton-based action recognition has been widely studied because skeleton data
convey compact information of body movement and have strong adaptability to
dynamic circumstances, e.g., variations in viewpoints, occlusions and
complicated background \cite{han2017space}. Previous works formulate skeleton data as a
sequence of grid-shaped joint-coordinate vectors and use CNNs \cite{HBRNN}\cite{ntu}\cite{ST-LSTM}\cite{STA-LSTM}\cite{VA-LSTM}\cite{li2018independently}\cite{cao2018skeleton} or recurrent
neural networks (RNNs) \cite{liu2017two}\cite{TCN}\cite{ke2017new}\cite{SynCNN}\cite{li2017skeleton}\cite{li2017skeleton2}\cite{li2017skeleton3} to learn the actions.
As skeleton data naturally lies in a non-Euclidean space with joints as vertexes and their connections in the human body as edges, CNN- and RNN-based methods cannot fully utilize the rich information conveyed in the graph structure of skeleton data.

Recently, graph convolutional networks, with their superior capability in dealing with graph data,
have been introduced to skeleton-based action recognition and have achieved
state-of-the-art performance \cite{Deep-STGCK}\cite{ST-GCN}\cite{DPRL}\cite{SR-TSL}\cite{STGR-GCN}\cite{AS-GCN}\cite{GR-GCN}\cite{2S-AGCN}\cite{NAS-GCN}.
Most of these methods focus on the design/learning of the graph topology.
Yan et al. \cite{ST-GCN} introduced GCNs to model skeleton data and constructed a predefined graph with a fixed topology
constraint.
Shi et al. \cite{2S-AGCN} proposed to learn an adaptive graph by
parameterizing the graphs, and then updated the graph jointly with convolutional
parameters.
Gao et al. \cite{GR-GCN} introduced a high-order approximation for
a larger receptive field of the graph.
Peng et al. \cite{NAS-GCN} tried to search for different
graphs at different layers via automatic neural architecture searching (NAS)
\cite{NAS}. However, once the graph is generated, only a single
scale and one transformation exist in each layer of the networks. As a consequence, the backbone of
these GCN-based methods has intrinsic limitations on extracting and
synthesizing information from different scales and transformations from
different paths at different levels.

Intuitively, many insights from the design philosophy of CNN
can be integrated into GCN-based backbone networks.
Specifically,
a) inputs can be split into different paths with a few lower-dimensional
embeddings or identity mappings (GoogLeNet \cite{GoogleNet}, ResNeXt
\cite{ResNeXt}, ResNet \cite{ResNet});
b) different sets of transformations can be applied to each path (GoogLeNet
\cite{GoogleNet}, ResNeXt \cite{ResNeXt});
c) the outputs of all the paths can be aggregated with concatenation
or summation (ResNet \cite{ResNet}, ResNeXt \cite{ResNeXt}, GoogLeNet
\cite{GoogleNet}, DenseNet \cite{DenseNet}).
However, due to the graph structure of skeleton data, it is very challenging to embed these insights in GCNs.
For example, for multi-scale spatial processing, instead of using multiple kernel sizes in CNNs,
a customized operation for different orders of hop connections is needed for GCNs.
For multi-view temporal processing (position and motion), motion features specified for
graph convolution of skeleton sequences are still not touched. In fact, many biological studies \cite{bio1}\cite{bio2}\cite{bio3}\cite{bio4}\cite{bio5} have shown that, 20\% of cells in primate visual systems are responsive to dynamic motion changes, but are not sensitive to spatial details \cite{slowfast}.

In this paper, we adopt the strategy of repeating layers in CNNs and reinvent the
split-transform-merge strategy in GCNs for spatial and temporal skeleton sequence processing in each layer.
For each layer, the inputs are split into three paths: a spatial path for spatial
features, a temporal path for sequential features, and a residual path for the
reuse of the input features, as shown in Fig. \ref{fig:fusion}.
The spatial path (named spatial inception) is further split into four branches.
We apply $1$st to $4$th order adjacency sampling as four sets of graph transformations with $1\times$ to
$4\times$ hop connections.
The following are the specified graph transformations
with $1\times1$ convolution, batch normalization and ReLU. The temporal path
(named temporal inception) consists of two sets of transformations.
One set is a direct graph convolution on position features of the same joints across consecutive frames,
and the other set is a graph convolution on motion features of the same joints across consecutive frames.
Notably, this is the first time that the motion features of joints have been used in skeleton-based action
recognition.
Finally, in the merging stage, the outputs of both spatial path and temporal path are
first concatenated and fused with $1\times1$ convolution.
Then, features of the three paths are aggregated by summation.
The whole block is named the spatio-temporal inception for
its analogy to inception modules in CNNs.

To verify the superiority of our proposed spatio-temporal inception graph
convolutional network
(STIGCN) for skeleton-based action recognition, extensive experiments are
conducted on two large-scale datasets.
Our network outperforms state-of-the-art methods by a significant margin with only $1/5$ of the
parameters and $1/10$ of the FLOPS. Furthermore, while other methods rely on a
two-stream pipeline that requires skeleton data and crafted bone data as
inputs, our method only requires raw skeleton data as input.

The contributions of this paper are summarized as follows:
\begin{itemize}
\item We propose a graph convolution backbone architecture, termed
    spatio-temporal inception graph convolutional network, for skeleton-based action recognition.
    This network overcomes the limitations of state-of-the-art methods in extracting and
    synthesizing information of different scales and transformations from
    different paths at different levels.

\item To overcome the gap of the convolution operation between CNNs and GCNs, we reinvent the split-transform-merge strategy in GCNs for skeleton sequence processing.

\item Our method indicates that increasing the number of transformation sets is a more effective way of gaining accuracy than simply creating wider GCNs. We hope this insight will facilitate the iteration of
    GCN-based backbones for spatio-temporal sequence analyses.

\item On two large-scale datasets for skeleton-based action recognition, the
    proposed network outperforms state-of-the-art methods by a significant margin with
    surprisingly fewer parameters and FLOPs. The code and pretrained models
    will be released to facilitate related future research.
\end{itemize}

\section{Related Work}
\label{sec:relatedWorks}
\subsection{Skeleton-Based Action Recognition}
In human action recognition, skeleton data have attracted increasing attention due to
their robustness against body scales, viewpoints and backgrounds. 
Conventional methods in skeleton-data-based human action recognition utilize handcrafted feature descriptors to model
the human body \cite{Lie-Group}\cite{fernando2015modeling}\cite{hussein2013human} \cite{weng2017spatio}.
However, these methods either ignore the information of interactions between specific
sets of body parts or suffer from complicated design processes.

CNN-based and RNN-based methods have been well investigated for skeleton-based action recognition.
CNN-based methods \cite{liu2017two}\cite{TCN}\cite{ke2017new}\cite{SynCNN}\cite{
li2017skeleton}\cite{li2017skeleton2}\cite{li2017skeleton3}
formulate skeleton data as a pseudo-image based on manually designed transformation rules.
RNN-based methods \cite{HBRNN}\cite{ntu}\cite{ST-LSTM}\cite{STA-LSTM}
\cite{VA-LSTM}\cite{li2018independently}\cite{cao2018skeleton}
focus on modeling the temporal dependency of the inputs, where joint data of the human body are
rearranged by grid-shaped structure.
However, both CNN-based and RNN-based models neglect the co-occurrence pattern between spatial and temporal features since the skeleton data are naturally embedded in the form of graphs rather than a vector sequence or 2D grid.

Recently, GCNs have been introduced to skeleton-based action recognition and have achieved
state-of-the-art performance.
Most of these methods focus on the design/learning of the graph topology. 
Yan et al. \cite{ST-GCN} first introduced GCNs to model skeleton data and constructed a predefined graph with a fixed topology
constraint. 
Shi et al. \cite{2S-AGCN} proposed learning an adaptive graph by
parameterizing the graphs and then updated the graph jointly with convolutional
parameters. 
Gao et al. \cite{GR-GCN} introduced a high-order approximation for
a larger receptive field of the graph and learned it by solving a sparsified
regression problem. 
Peng et al. \cite{NAS-GCN} tried to search for different
graphs at different layers via NAS \cite{NAS}.

\subsection{Backbone Convolutional Neural Networks}
Many works have shown that synthesizing the outputs of different information paths in
a building block is helpful.
Deep neural decision forests \cite{kontschieder2015deep} are tree-patterned
multi-branch networks with learned splitting functions.
GoogLeNet \cite{GoogleNet} uses an inception module to introduce multi-scale
processing in different paths of the building block. The generated multi-scale features
are merged by concatenation.
ResNet \cite{ResNet} uses a residual learning framework in which the identity
mapping of the inputs and the convolutional outputs are merged through elementwise
addition.
ResNeXt \cite{ResNeXt} designs a building block that aggregates a set of transformations.
In DenseNet \cite{DenseNet}, the feature maps of all the preceding layers
are fed into the current layer, and the feature maps of this layer are used as
inputs to all the subsequent layers. A transition layer is designed to synthesize
the feature maps of all the layers in one dense block.
Qiu et al. \cite{p3d} split 3$\times$3$\times$3 convolutions into 1$\times$3$\times$3 convolutional
filters on spatial domain and 3$\times$1$\times$1 convolutions on temporal
connections between adjacent feature maps.
LocalCNN \cite{localcnn} uses a local operation as genetic building blocks for synthesizing global and local information in any layer. In
the local path, Yang et al. \cite{localcnn} used a sampling module to extract local regions from
the inputs, and the feature extraction module and feature fusion module were
designed to transform and merge features.

\subsection{Graph Convolutional Networks}
GCNs are widely used on irregular data, e.g., social networks and
biological data. The key challenge is to define convolutions over graphs, which
is difficult due to the unordered graph data. The principle of constructing GCNs
mainly follows the spatial perspective or the spectral perspective.
Spatial perspective methods \cite{duvenaud2015convolutional}\cite{niepert2016learning}\cite{
hamilton2017inductive}\cite{monti2017geometric}\cite{kipf2018neural}
directly perform convolutions on the graph vertexes
and their neighbors, then normalize the outputs based on manually designed rules.
Spectral GCNs transform graph signals into spectral domains by graph
Laplacian methods \cite{duvenaud2015convolutional}\cite{henaff2015deep},
and then apply spectral filters on the spectral domains. In \cite{hammond2011wavelets}, Chebyshev expansions are used to approximate the graph Fourier transform, and the graph convolution is well approximated by a weighted summation of Chebyshev transformations over the skeleton data.

\section{Approach}
\label{sec:approach}

\begin{figure*}[h]
  \centering
  \includegraphics[width=\linewidth]{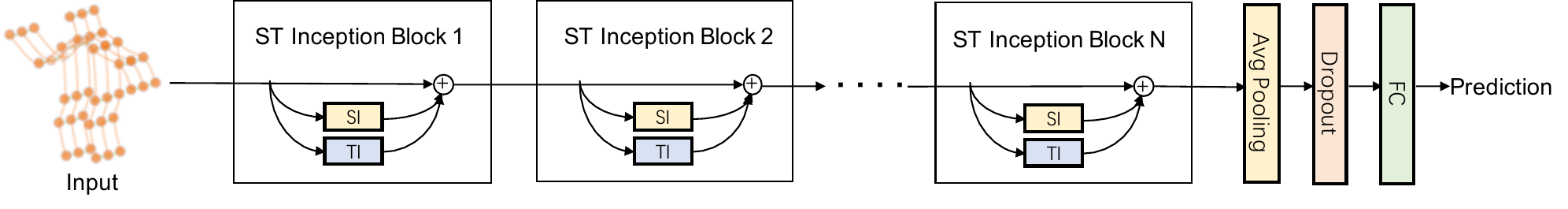}
  \caption{The overall architecture of our spatio-temporal inception graph
      convolutional network. It consists of a stack of ST inception blocks shown
  in Fig. \ref{fig:fusion} (b).
      This network takes raw skeleton data as inputs and is trained in an
  end-to-end manner.}
  \label{fig:framework}
\end{figure*}


\subsection{Motivation}
The topology of the adjacency graph is the key factor for modeling correlations of the
input skeletons. Therefore, state-of-the-art methods, including NAS \cite{NAS-GCN}, adaptive
graph learning \cite{2S-AGCN} and sparsified graph regression \cite{GR-GCN},  mainly focus on the design/learning of
the graph topology. 
However, once the graph is generated, only a single scale and one transformation exist in each layer of the networks. 
As a consequence, the backbone of these methods has intrinsic limitations on extracting and synthesizing information of different scales and transformations from different paths at different levels.

Intuitively, the success of the split-transform-merge strategy in recently developed backbone
convolutional neural networks could be adopted for GCN-based backbone networks.
However, due to the gap between graph skeleton data and traditional images/videos, it is not trivial to apply the split-transform-merge strategy in CNNs to GCNs.
Specified modules are required to extract and synthesize features from multiple scales and transformations on graph data.
To solve this problem, we design such modules, including a multi-scale spatial graph convolution module and a motion graph convolution module, and propose a simple graph convolution backbone architecture for skeleton-based action recognition.


\subsection{Instantiation}
\label{sec:instantiation}
In this section, we describe our design of spatio-temporal inception block
for skeleton-based action recognition. First, we briefly show how to construct a multi-scale spatial
graph convolution.

Consider an undirected graph $\mathcal{G}=\left\{\mathcal{V}, \mathcal{E}, A\right\}$
composed of $n=|\mathcal{V}|$ nodes. The nodes are connected by $|\mathcal{E}|$
edges and the connections are encoded in the adjacency matrix
$A\in\mathcal{R}^{n\times n}$. $F_{in}\in\mathcal{R}^n$ is the input
representation of $\mathcal{G}$. After a graph Fourier transform, the convolutional filtering in spatial
domain could be formulated as an inner-product operation in spectral domain \cite{cheby}.
Specifically, the graph Laplacian $L$, of which the normalized
definition is $L=I_n-D^{-1/2}AD^{-1/2}$ and $D_{ij}=\sum_j A_{ij}$, is used for
Fourier transform. Then a graph filtered by operator $g_\theta$, parameterized
by $\theta$, can be formulated as
\begin{equation}
    F_{out}=g_\theta(L)F_{in}=Ug_\theta(\Lambda)U^TF_{in},
\end{equation}
where $F_{out}$ is the output feature of the input graph, $U$ is the Fourier basis,
$L=U\Lambda U^T$, and $\Lambda$ is the corresponding eigenvalue of $L$.
Hammond et al. \cite{cheby} proved that the filter $g_\theta$ could be well approximated by
$R$th order Chebyshev polynomials,
\begin{equation}\label{eq:combine}
    F_{out}=\sum_{r=0}^{R} \theta_r^\prime T_r(\hat{L})F_{in},
\end{equation}
where $\theta_r^\prime$ denote Chebyshev coefficients. The Chebyshev polynomial
is recursively defined as
\begin{equation}\label{eq:cheby}
    T_r(\hat{L})=2\hat{L}T_{r-1}(\hat{L})-T_{r-2}(\hat{L}')
\end{equation}
with $T_0=1$ and $T_1=\hat{L}$. $\hat{L}=2L/\lambda_{max}-I_n$ is
normalized to $[-1, 1]$.

In general, the graph $L$ filtered by $g_\theta$ can be approximated as a linear
combination of input representation transformed by Chebyshev polynomials.
\emph{As a consequence, a spatial graph convolution with a receptive field of $k$ can be
formulated as a linear transformation of a $k$th order Chebyshev polynomial matrix.}

\subsubsection{Spatial Inception}
An overall instantiation of our ST inception building block is shown in
Fig. \ref{fig:fusion} (b). It consists of a spatial-inception (SI) path,
a temporal inception (TI) path and a residual path.
There are three components in each inception path: sampling, convolution
and fusion. We now present the details of spatial inception first.

\textbf{Adjacency Sampling.}
Inspired by the spectral formulation of graph convolutions, we reformulate
the feature sampling module as a matrix multiplication operation between skeleton
representations and a graph transformation defined as Chebyshev polynomials.
The $r$th order Chebyshev polynomial $T_r(\hat{L})$, as defined in
Eq. (\ref{eq:cheby}), represents the $r$th order hop connections between
skeleton joints.
Fig. \ref{fig:dataset} shows that most joints in the graph can be reached by
4 hop connections from the center joint (labeled as $1$).
Thus we choose $R=4$ to approximate the multi-scale graph filtering operation.
A higher order approximation may bring larger performance gains with additional computational
costs, but this direction is not the priority of this paper.
In detail, the $0$th to $4$th order Chebyshev polynomials are defined as follows:
\begin{equation}
    \begin{aligned}
    &T_0=I,\\
    &T_1=\hat{L},\\
    &T_2=2\hat{L}^2-I,\\
    &T_3=4\hat{L}^3-3\hat{L},\\
    &T_4=8\hat{L}^4-8\hat{L}^2+I.
    \end{aligned}
\end{equation}

As illustrated in Fig. \ref{fig:fusion}, there are $4$ branches in the SI path, 
corresponding to the 1st to 4th order graph transformations, respectively.
$T_0$ represents the identity transformation, which is identical to the residual connection.
Therefore, the $0$th order sampling module is already included in the residual path.
Following the adaptive graph topology introduced in \cite{2S-AGCN},
we apply layer-dependent bias and data-dependent bias to the
transformation matrix for more flexible hop connections.
Other parameters of the predefined transformation matrix are fixed
during training except for the adaptive bias.

\textbf{Convolution Module.}
The graph convolution module is used to extract graph features of every scale.
It consists of a $1\times1$ convolutional layer, a batch
normalization layer and a ReLU layer. The number of output feature maps is set
to $1/4$ of the total width of the spatial path for computational efficiency.
Bottleneck-like architecture will be investigated in our future work.

\textbf{Fusion Module.}
The feature fusion module is introduced to generate more robust and discriminative
representations by synthesizing outputs of all paths. In this
paper, the feature fusion module is formed as a concatenation layer of all the
outputs, followed by a $1\times1$ convolutional layer with batch
normalization and ReLU. The number of output channels of the $1\times1$ convolutional
layer is set to the number of input channels to maintain the cardinality.

\subsubsection{Temporal Inception}
As shown in Fig. \ref{fig:fusion} (b), there are two branches in TI path.
One branch directly takes features of the same joints in consecutive frames as inputs for
position feature processing. 
The other branch feeds inputs into the motion
sampling module for motion feature processing. 
This is the first time that motion features of joints are used in skeleton-based action recognition.

\textbf{Motion Sampling.}
The second-order spatial information, i.e., the bone information, was first introduced in \cite{2S-AGCN} and then widely used in later works
\cite{NAS-GCN}\cite{GR-GCN}. 
However, the second-order temporal information is still ignored in previous works. 
In this paper, we design a motion sampling module
to explicitly model the second-order temporal information, termed the motion information.
In particular, the motion information is defined as the difference between
consecutive frames. For example, given a frame of skeleton data at time $t$,
\begin{equation*}
    \begin{split}
    v_t=\{(x_1^{(t)}, y_1^{(t)}, z_1^{(t)}),
    \cdots, (x_n^{(t)}, y_n^{(t)}, z_n^{(t)})\}
    \end{split}
\end{equation*}
where $(x_i^{(t)}, y_i^{(t)}, z_i^{(t)})$ are the 3D coordinates of the $i$th
joint at time $t$ and its next frame at time $t+1$ is
\begin{equation*}
    \begin{aligned}
    v_{t+1}=\{(x_1^{(t+1)}, y_1^{(t+1)}, z_1^{(t+1)}),
    \cdots, (x_n^{(t+1)}, y_n^{(t+1)}, z_n^{(t+1)})\}.
    \end{aligned}
\end{equation*}
The vector of the motion is calculated as
\begin{equation*}
    \begin{aligned}
    m_t=v_{t+1}-v_t=
    \{&(x_1^{(t+1)}-x_1^{(t)}, y_1^{(t+1)}-y_1^{(t)}, z_1^{(t+1)}-z_1^{(t)}),
        \cdots,\\
    &(x_n^{(t+1)}-x_n^{(t)}, y_n^{(t+1)}-y_n^{(t)}, z_n^{(t+1)}-z_n^{(t)})\}.
    \end{aligned}
\end{equation*}

The motion information can also be considered the optical flow of skeleton sequence.
The joints of the skeleton data are similar to observed objects in RGB videos, and the optical flow is calculated as the relative motion of objects between consecutive frames.
Therefore, it is natural to utilize the aforementioned
motion sampling operation for motion feature processing.

\textbf{Convolution and Fusion.}
The feature extraction module is designed to extract features from the frame
sequence and motion sequence.
Different from the convolution in the SI path, we use a $3\times1$
kernel for temporal convolution, where kernel size $3$ corresponds to the
temporal span, to construct temporal connections on adjacent feature maps in the
input sequence.
The feature fusion module concatenates the outputs of the two temporal branches,
followed by a $1\times1$ convolution, batch normalization and ReLU.

\subsubsection{Spatio-Temporal Fusion}
In the final merging stage, the outputs of spatial path, temporal path and
residual path are aggregated by summation.

\subsection{Network Architecture}
To maintain consistent with state-of-the-art GCNs
\cite{ST-GCN}\cite{2S-AGCN}\cite{NAS-GCN}, we introduce ten ST inception blocks
into our STIGCN.
The overall architecture is
illustrated in Fig. \ref{fig:framework} and Table \ref{tab:arch}. It is a stack
of basic building blocks shown in Fig. \ref{fig:fusion} (b).
There are four stages in STIGCN, consisting of $1$, $3$, $3$ and $3$ building blocks,
respectively. The numbers of output channels for these blocks are $64$, $64$,
$64$, $64$, $128$, $128$, $128$, $256$, $256$ and $256$, respectively. Inside each block,
``S=4'' and ``T=2'' denote the numbers of branches in SI and TI, respectively,
followed by the size of kernels in convolution modules and fusion modules.
A batch normalization layer is added to the beginning to normalize the input data.
Max pooling is applied after the first three stages, to construct a temporal hierarchical structure. The extracted features of the last block are fed into a global average pooling layer to pool feature maps of different samples to the same size. After a dropout layer, a softmax classifier is used to generate the final prediction.

\begin{table}
  \begin{tabular}{c|c|cc}
    \hline
    layer name & output size & components \\ \hline
     & $3\times300\times N_j$ & data batch normalization \\
    \hline
    Stage 1 & $64\times300\times N_j$ & $\begin{bmatrix} S=4 & T=2 \\ 1\times1, 16 &
    1\times3, 32 \\ 1\times1, 64 & 1\times1, 64 \end{bmatrix}\times1$ \\
    \hline
    \multirow{4}{*}{\centering Stage 2} & \multirow{4}{*}{$64\times150\times N_j$} & $1\times2$ max pooling, stride $1\times2$ \\ \cline{3-3}
     & & $\begin{bmatrix} S=4 & T=2 \\ 1\times1, 16 & 1\times3, 32 \\1\times1, 64 & 1\times1, 64 \end{bmatrix}\times3$ \\
    \hline
    \multirow{4}{*}{\centering Stage 3} & \multirow{4}{*}{$128\times75\times N_j$} & $1\times2$ max pooling, stride $1\times2$ \\ \cline{3-3}
     & & $\begin{bmatrix} S=4 & T=2 \\ 1\times1, 32 & 1\times3, 64 \\1\times1, 128 & 1\times1, 128 \end{bmatrix}\times3$ \\
    \hline
    \multirow{4}{*}{\centering Stage 4} & \multirow{4}{*}{$256\times37\times N_j$} & $1\times2$ max pooling, stride $1\times2$ \\ \cline{3-3}
     & & $\begin{bmatrix} S=4 & T=2 \\ 1\times1, 64 & 1\times3, 128 \\1\times1, 256 & 1\times1, 256 \end{bmatrix}\times3$ \\
    \hline
     & 256 & avg pooling, dropout \\
    \hline
    classifier & $N_c$ & fc \\
    \hline
  \end{tabular}
  \caption{Architecture of spatio-temporal inception graph convolutional networks. 
      ``S=4'' and ``T=2'' denote the numbers of branches in spatial inception and temporal inception, respectively, followed by the size of kernels in convolution modules and fusion modules.
      $N_j$ is the number of joints in the graph. $N_c$ is the number of action classes.}
\label{tab:arch}
\vspace{-7mm}
\end{table}

\section{Experiments}
\label{sec:experiments}
\subsection{Datasets and Evaluation Protocol} \label{sec:dataset}
\textbf{NTU RGB+D} \cite{ntu} is the most widely used and the largest
multi-modality indoor-captured action recognition dataset. It contains
$56,880$ action clips (samples) from $60$ action classes. 
For classification task, we follow the benchmark evaluations in the original
work \cite{ntu}, which are cross-subject (X-Sub) and cross-view (X-View)
evaluations. In X-Sub evaluation, $40,320$ samples performed by $20$ subjects
are used as the training set, while the rest belong to the testing set.
X-View evaluation divides the dataset according to camera views, where
training and testing sets have $37,920$ and $18,960$ samples, respectively.
For retrieval task, we follow the settings in \cite{ntu120} and split the dataset
into two parts: a training set containing $47,180$ samples from
50 action classes and a testing set containing $9,700$ samples from the remaining 10
action classes. No data augmentation is performed in either task, and the data processing procedure is the
same as which in \cite{ST-GCN}.

\begin{figure}[!t]
  \centering
  \includegraphics[width=0.55\linewidth]{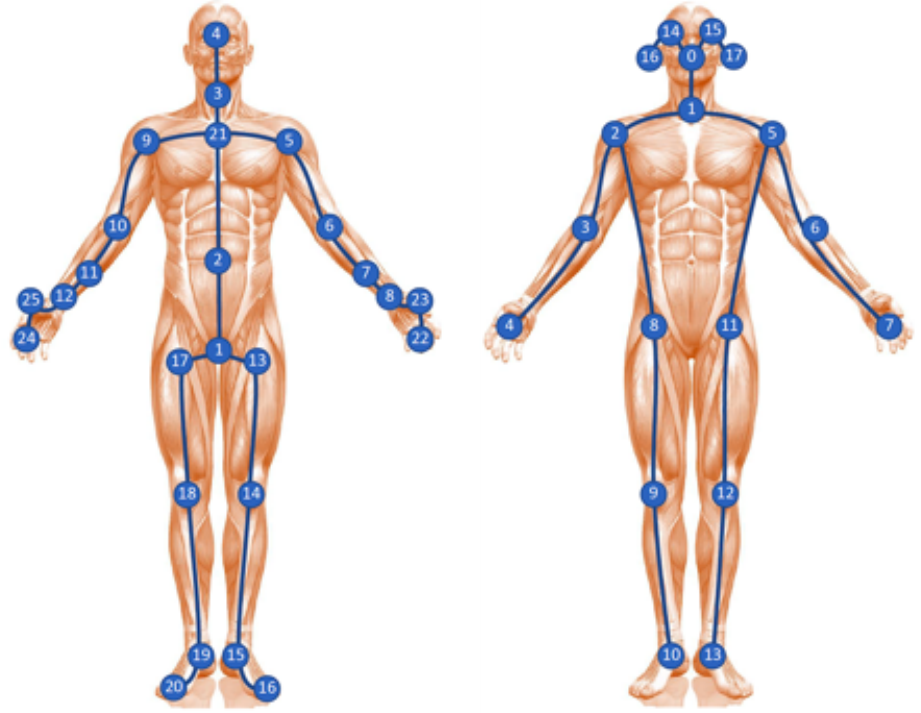}
  \vspace{-3mm}
  \caption{Left: the 25 joint labels in NTU RGB+D. Right: the 18 joint labels in Kinetics-Skeleton.}
  \label{fig:dataset}
  \vspace{-5mm}
\end{figure}

\textbf{Kinetics-Skeleton} \cite{kinetics} is a large-scale human action dataset
that contains $260,000$ video clips from $400$ action classes. Yan
et al. \cite{ST-GCN} employed the open source toolbox OpenPose \cite{openpose} to estimate coordinates of 18 joints in each frame.
For classification task, the dataset is divided into a training set ($240,000$
samples) and a testing set ($20,000$ samples).
For retrieval task, the dataset is randomly divided into a training set 
containing $228,273$ samples from
350 classes and a testing set containing $31,727$ samples from the remaining 50
action classes.
We use the same data augmentation as in \cite{ST-GCN}. 
The definitions of the joints and their natural connections in these two datasets are
shown in Fig. \ref{fig:dataset}.

\textbf{Evaluation Protocol}.
We calculate the top-1
accuracy on NTU RGB+D and the top-1/top-5 accuracy on Kinetics-Skeleton
to evaluate the performance. And in retrieval task, we calculate the mean average precision (mAP) and cumulative
matching characteristics (CMC) at rank-1 on both datasets to evaluate the performance.

\subsection{Implementation Details}
Our framework is implemented on PyTorch \cite{pytorch} and the code will be
released later.
Following \cite{2S-AGCN}, all experiments use stochastic gradient descent with a Nesterov momentum of $0.9$.
For NTU RGB+D, the batch size is $64$, the weight decay is $5e-4$ and
the initial learning rate is $0.1$. The learning rate is divided by $10$ at the
$30$th and $40$th epochs. The training process ends at the $50$th epoch.
For Kinetics-Skeleton, the batch size is $128$ and the training lasts
$60$ epochs. The learning rate is set to $0.1$ at the beginning and is divided by
$10$ at the $45$th and $50$th epochs. The weight decay is $1.5e-4$.

\subsection{Comparison with State-of-the-Art Methods}
\begin{table}
  \begin{tabular}{c|c|cc}
    \toprule
    Input & Method & X-Sub(\%) & X-View(\%) \\
    \midrule
    \multirow{18}{*}{\centering Joint} & Lie Group \cite{Lie-Group} & 50.1 & 82.8 \\
    \cline{2-4}
    & HBRNN \cite{HBRNN} & 59.1 & 64.0 \\
    & Deep LSTM \cite{ntu} & 60.7 & 67.3 \\
    & P-LSTM \cite{ntu} & 62.9 & 70.3 \\
    & ST-LSTM \cite{ST-LSTM} & 69.2 & 77.7 \\
    & STA-LSTM \cite{STA-LSTM} & 73.4 & 81.2 \\
    & VA-LSTM \cite{VA-LSTM} & 79.2 & 87.7 \\
    \cline{2-4}
    & TCN \cite{TCN} & 74.3 & 83.1 \\
    & SynCNN \cite{SynCNN} & 80.0 & 87.2 \\
    \cline{2-4}
    & Deep STGCK \cite{Deep-STGCK} & 74.9 & 86.3 \\
    & ST-GCN \cite{ST-GCN} & 81.5 & 88.3 \\
    & DPRL \cite{DPRL} & 83.5 & 89.8 \\
    & SR-TSL \cite{SR-TSL} & 84.8 & 92.4 \\
    & STGR-GCN \cite{STGR-GCN} & 86.9 & 92.3 \\
    & AS-GCN \cite{AS-GCN} & 86.8 & 94.2 \\
    & GR-GCN \cite{GR-GCN} & 87.5 & 94.3 \\
    & 2S-AGCN \cite{2S-AGCN} & 86.6 & 93.7  \\
    & NAS-GCN \cite{NAS-GCN} & 87.6  & 94.5  \\
    & \textbf{STIGCN (ours)} & \textbf{90.1} & \textbf{96.1} \\
    \hline
    \multirow{2}{*}{\centering Joint+Bone} & 2S-AGCN \cite{2S-AGCN} & 88.5 & 95.1  \\
    & NAS-GCN \cite{NAS-GCN} & 89.4 & 95.7  \\
  \bottomrule
\end{tabular}
\caption{Comparison of classification accuracy on NTU RGB+D.}
\label{tab:ntu_cls}
\vspace{-10mm}
\end{table}
\subsubsection{Action Classification}


Our method is compared to state-of-the-art methods, including handcrafted-feature-based methods \cite{Lie-Group},
RNN-based methods
\cite{HBRNN}\cite{ntu}\cite{ST-LSTM}\cite{STA-LSTM}\cite{VA-LSTM},
CNN-based methods \cite{TCN}\cite{SynCNN},
and GCN-based methods
\cite{Deep-STGCK}\cite{ST-GCN}\cite{DPRL}\cite{SR-TSL}\cite{STGR-GCN}\cite{AS-GCN}\cite{GR-GCN}\cite{2S-AGCN}\cite{NAS-GCN}.
The results on NTU RGB+D and Kinetics-Skeleton are summarized in Table \ref{tab:ntu_cls} and
Table \ref{tab:kinetics_cls}, respectively.
We can see that STIGCN outperforms other methods on both datasets by a notable margin .

It is worth noting that both 2S-AGCN and NAS-GCN use extra bone data.
They first train two independent models with joint data and bone data respectively, then ensemble the
outputs of them during testing. STIGCN is trained in an
end-to-end manner and outperforms aforementioned methods without any ensemble or
extra bone data. 
When only joint data are used in 2S-AGCN and NAS-GCN, STIGCN outperforms them by $3.5\%$ and $2.5\%$ on NTU RGB+D X-Sub.
We can conclude that STIGCN is much better at leveraging the multi-scale and multi-view knowledge
from the joint sequence data, which significantly boosts the performance.

Moreover, Table \ref{tab:param} illustrates that STIGCN needs much fewer
parameters and FLOPs than state-of-the-art methods. It is $1/8$
parameters and $1/18$ FLOPs compared with NAS-GCN \cite{NAS-GCN}. This finding shows that
STIGCN is more efficient at extracting representations from the
skeleton sequence, which is important in practical scenarios. More importantly,
as the width and depth of our network is the same as the network for single-stream
inputs in 2S-GAN \cite{2S-AGCN} and NAS-GAN \cite{NAS-GCN}, the superior
performance indicates that increasing
the number of transformation sets is a more effective way of gaining accuracy
than simply creating wider GCNs.

\begin{figure}[!t]
  \centering
  \includegraphics[width=0.8\linewidth]{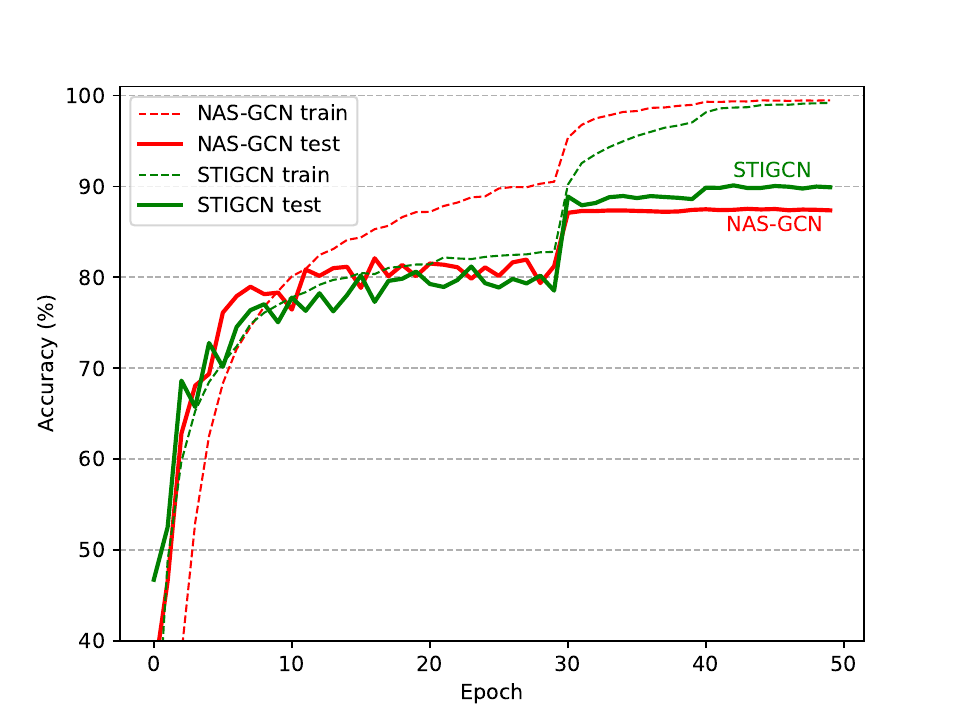}
  \vspace{-5mm}
  \caption{Training and testing curves on NTU RGB+D X-Sub. 
  }
  \label{fig:loss}
\end{figure}

Fig. \ref{fig:loss} shows the training and testing curves of STIGCN and NAS-GCN
(joint) on NTU RGB+D X-Sub. STIGCN exhibits lower training accuracy
but higher testing accuracy, which indicates that STIGCN is more generalizable to
the testing data. With much fewer parameters and FLOPs, STIGCN achieves a higher generalization
capacity by alleviating the overfitting problem.

\begin{table}
  \begin{tabular}{c|c|cc}
    \toprule
    Input & Method & Top-1(\%) & Top-5(\%) \\
    \midrule
    \multirow{7}{*}{\centering Joint} & Feature \cite{fernando2015modeling} & 14.9 & 25.8 \\
    & P-LSTM \cite{ntu} & 16.4 & 35.3 \\
    & TCN \cite{TCN} & 20.3 & 40.0 \\
    & ST-GCN \cite{ST-GCN} & 30.7 & 52.8 \\
    & AS-GCN \cite{AS-GCN} & 34.8 & 56.5 \\
    & 2S-AGCN \cite{2S-AGCN} & 35.1 & 57.1  \\
    & NAS-GCN \cite{NAS-GCN} & 35.5  & 57.9 \\
    & \textbf{STIGCN (ours)} & \textbf{37.9} & \textbf{60.8} \\
    \hline
    \multirow{2}{*}{\centering Joint+Bone} & 2S-AGCN \cite{2S-AGCN} & 36.1 & 58.7  \\
    & NAS-GCN \cite{NAS-GCN} & 37.1  & 60.1 \\
  \bottomrule
\end{tabular}
\caption{Comparison of classification accuracy on Kinetics-Skeleton.}
\label{tab:kinetics_cls}
\vspace{-10mm}
\end{table}

\begin{table}
  \begin{tabular}{c|cc}
    \toprule
    Method & Params(M) & GFLOPs \\
    \midrule
    2S-AGCN \cite{2S-AGCN} & 7.0 & 37.3 \\
    NAS-GCN \cite{NAS-GCN} & 13.0 & 73.2\\
    \textbf{STIGCN (ours)} & \textbf{1.6} & \textbf{4.0} \\
  \bottomrule
\end{tabular}
\caption{Comparison of number of parameters and FLOPs.}
\label{tab:param}
\vspace{-10mm}
\end{table}

\subsubsection{Action Retrieval}
To further validate the representation learning capability of STIGCN,
we validate its performance on action retrieval tasks. Following the
configurations in \cite{ntu120}, we split the dataset into training
and testing sets without action class overlap. The model is trained on
training set with only the cross entropy loss, and tested by single query among all
testing samples.

We choose 2S-AGCN \cite{2S-AGCN} and NAS-AGCN \cite{NAS-GCN}
as baseline methods and train them by replicating the same training hyperparameters
and architectures as those in the original papers.
Features extracted from the trained models are used for retrieval.
Since both methods are trained with a two-stream pipeline, we train two models using joint-skeleton data and
bone-skeleton data separately and concatenate the output features for similarity calculation during retrieval. Our STIGCN uses only single-stream joint-skeleton data.
For all three methods, outputs of the final average pooling layer are used
as features for retrieval.

Table \ref{tab:ntu_ret} and Table \ref{tab:kinetics_ret} demonstrate that STIGCN
achieves the best performance.
Specifically, compared with NAS-GCN \cite{NAS-GCN},
our model achieves a 2.04\% mAP gain on NTU RGB+D and a 1.04\% gain on Kinetics-Skeleton
with only $1/8$ of the parameters, $1/18$ of the FLOPs and $1/2$ of the output features.
When the feature dimension is the same, the gains become 6.04\% and 2.05\%. The
superior performance of STIGCN in the retrieval task reveals that the proposed
GCN backbone is better at general representation learning for skeleton
sequence data.

\begin{table}
  \begin{tabular}{c|c|c|cc}
    \toprule
    Input & Method & Feat dim & mAP(\%) & CMC(\%) \\
    \midrule
    \multirow{2}{*}{\centering Joint} & 2S-AGCN \cite{2S-AGCN} & 256 & 73.83 & 93.97 \\
    & NAS-GCN \cite{NAS-GCN} & 256 & 74.04 & 94.12 \\
    & \textbf{STIGCN (ours)} & \textbf{256} & \textbf{80.08} & \textbf{96.17} \\
    \hline
    \multirow{2}{*}{\centering Joint+Bone} & 2S-AGCN \cite{2S-AGCN} & 512 & 77.18 & 95.36 \\
    & NAS-GCN \cite{NAS-GCN} & 512 & 78.04 & 95.78 \\
  \bottomrule
\end{tabular}
\caption{Comparison of action retrieval results on NTU RGB+D. ``Feat dim'': the feature dimension.}
\label{tab:ntu_ret}
\vspace{-5mm}
\end{table}

\begin{table}
  \begin{tabular}{c|c|c|cc}
    \toprule
    Input & Method & Feat dim & mAP(\%) & CMC(\%) \\
    \midrule
    \multirow{2}{*}{\centering Joint} & 2S-AGCN \cite{2S-AGCN} & 256 & 15.51 & 41.01  \\
    & NAS-GCN \cite{NAS-GCN} & 256 & 16.13 & 41.88 \\
    & \textbf{STIGCN (ours)} & \textbf{256} & \textbf{18.18} & \textbf{44.35} \\
    \hline
    \multirow{2}{*}{\centering Joint+Bone} & 2S-AGCN \cite{2S-AGCN} & 512 & 16.23 & 42.30  \\
    & NAS-GCN \cite{NAS-GCN} & 512 & 17.04 & 43.07 \\
  \bottomrule
\end{tabular}
\caption{Comparison of action retrieval result on Kinetics-Skeleton.}
\label{tab:kinetics_ret}
\vspace{-7mm}
\end{table}

\subsection{Ablation Analysis}

\subsubsection{Architecture.}
To validate the effectiveness of the proposed transformations in spatio-temporal inception block, including adjacency sampling of
multiple orders, motion sampling and feature fusion, we present the performance
of models with and without these components on NTU RGB+D X-Sub. To ensure a fair
comparison, the number of channels inside the block is fixed across all settings.
For example, if the number of channels of spatial path in ($a$) is $n$, then the number of channels in each of the two
branches in spatial path in ($b$) is $n/2$. 
If the number of channels of temporal path in ($a-e$) is $m$, then the number of channels in each of the two branches of temporal path in ($f$) is $m/2$.

Table \ref{tab:component} shows that the proposed network consistently
benefits from the introduced transformations. Moreover, settings ($e$) and ($f$) show
that the motion sampling module and feature fusion module provide much help in action recognition due to the synthesis of
motion information and multiple scale information.

\begin{table*}
\begin{tabular}{c|cccccc|c}
    \toprule
    Setting & Order=1 & Order=2 & Order=3 & Order=4 & Motion & Fusion & Acc (\%) \\
    \midrule
    a &\checkmark&&&&&& 86.53 \\
    b &\checkmark&\checkmark&&&&& 87.67 \\
    c &\checkmark&\checkmark&\checkmark&&&& 88.22 \\
    d &\checkmark&\checkmark&\checkmark&\checkmark&&& 88.49 \\
    e &\checkmark&\checkmark&\checkmark&\checkmark&\checkmark&& 89.45 \\
    f &\checkmark&\checkmark&\checkmark&\checkmark&\checkmark&\checkmark& 90.10 \\
    \bottomrule
\end{tabular}
\caption{Performance of different settings in STIGCN. ``Order=$i$''
means that there are $i$ adjacency sampling modules in spatial path.
``Motion'' denotes the motion sampling module, and ``Fusion'' represents the feature fusion module.
}
\label{tab:component}
\vspace{-5mm}
\end{table*}

\subsubsection{Fusion of Multiple Order Information.}
\begin{figure}[!t]
  \centering
  \subfigure[SI weight value]
  {
      \includegraphics[width=0.45\linewidth]{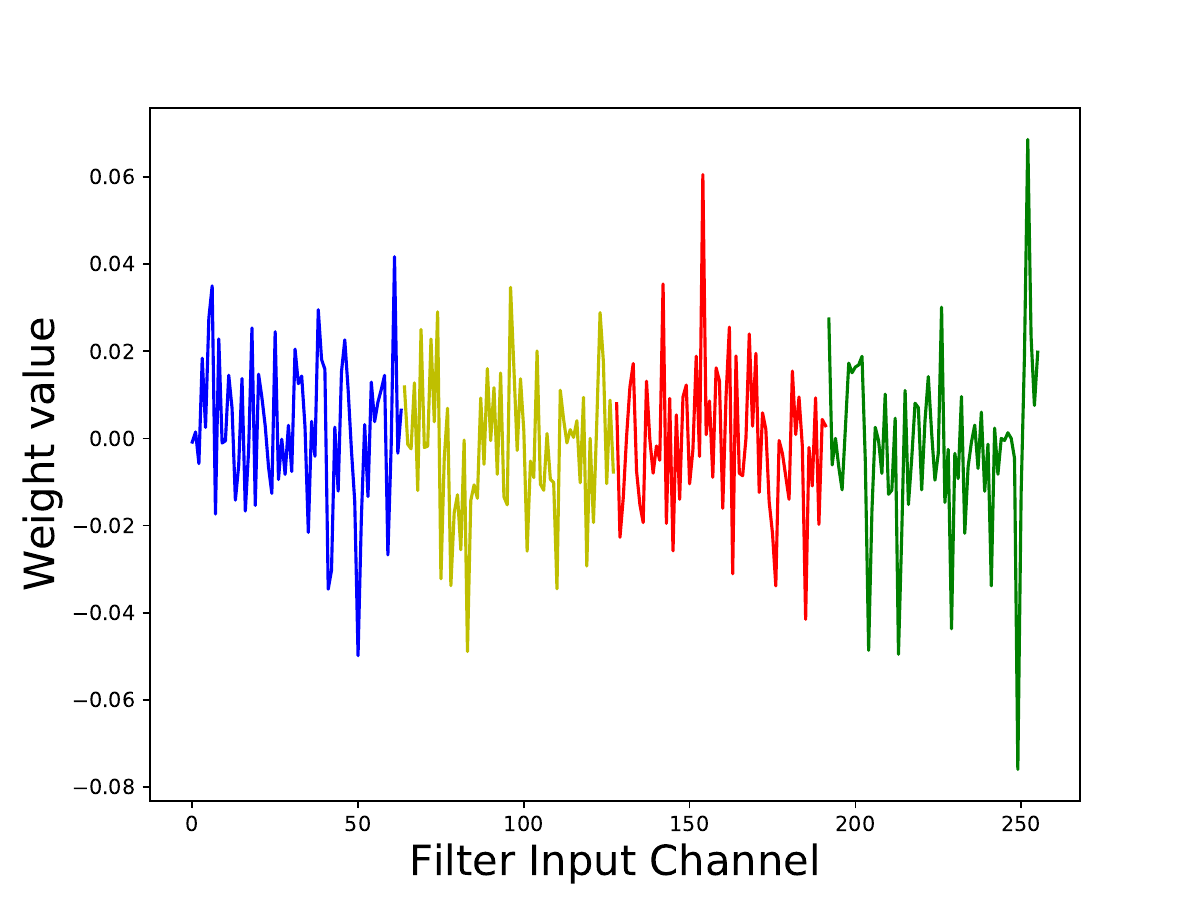}
  }
  \subfigure[TI weight value]
  {
      \includegraphics[width=0.45\linewidth]{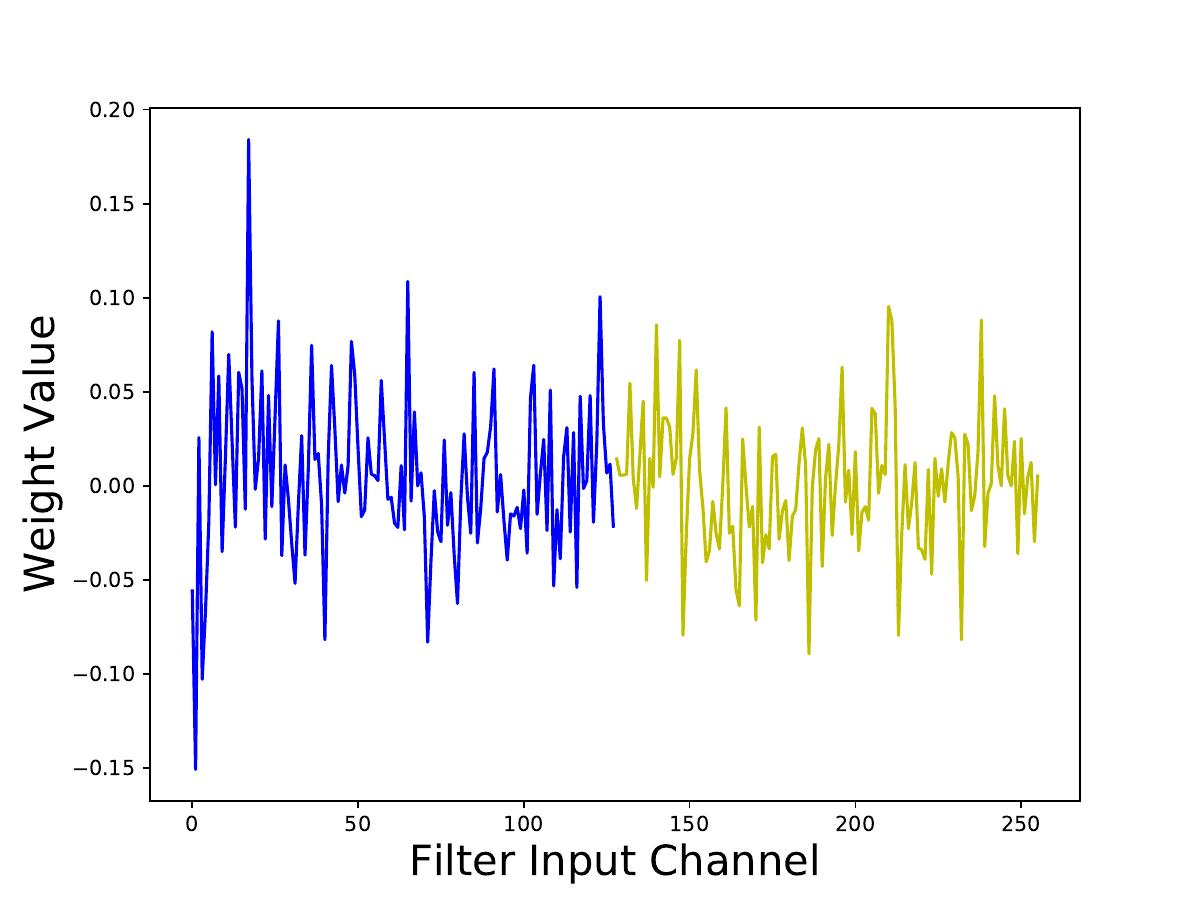}
  }
  \vspace{-3mm}
  \subfigure[SI input feature]
  {
      \includegraphics[width=0.6\linewidth]{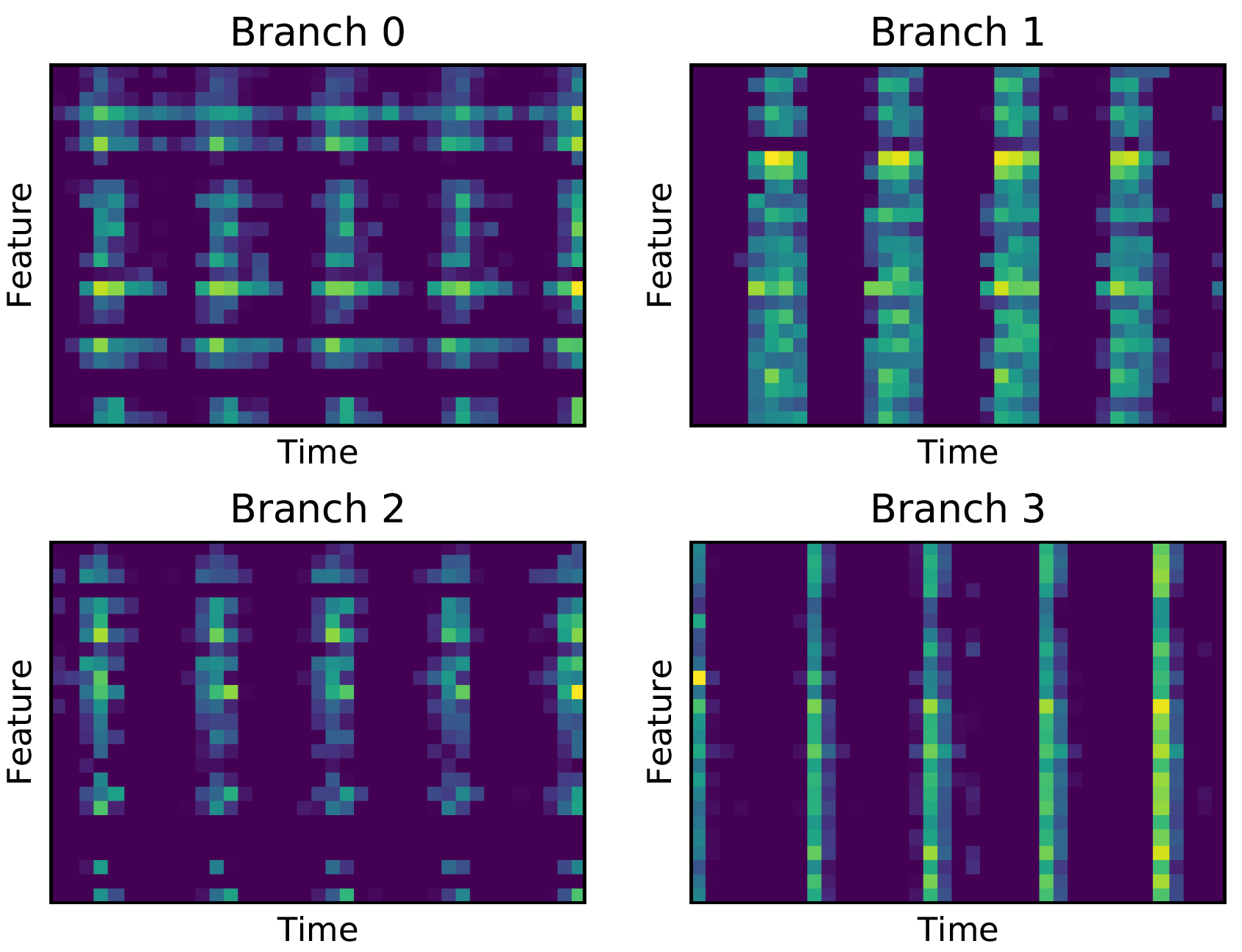}
  }
  \subfigure[TI input feature]
  {
      \includegraphics[width=0.295\linewidth]{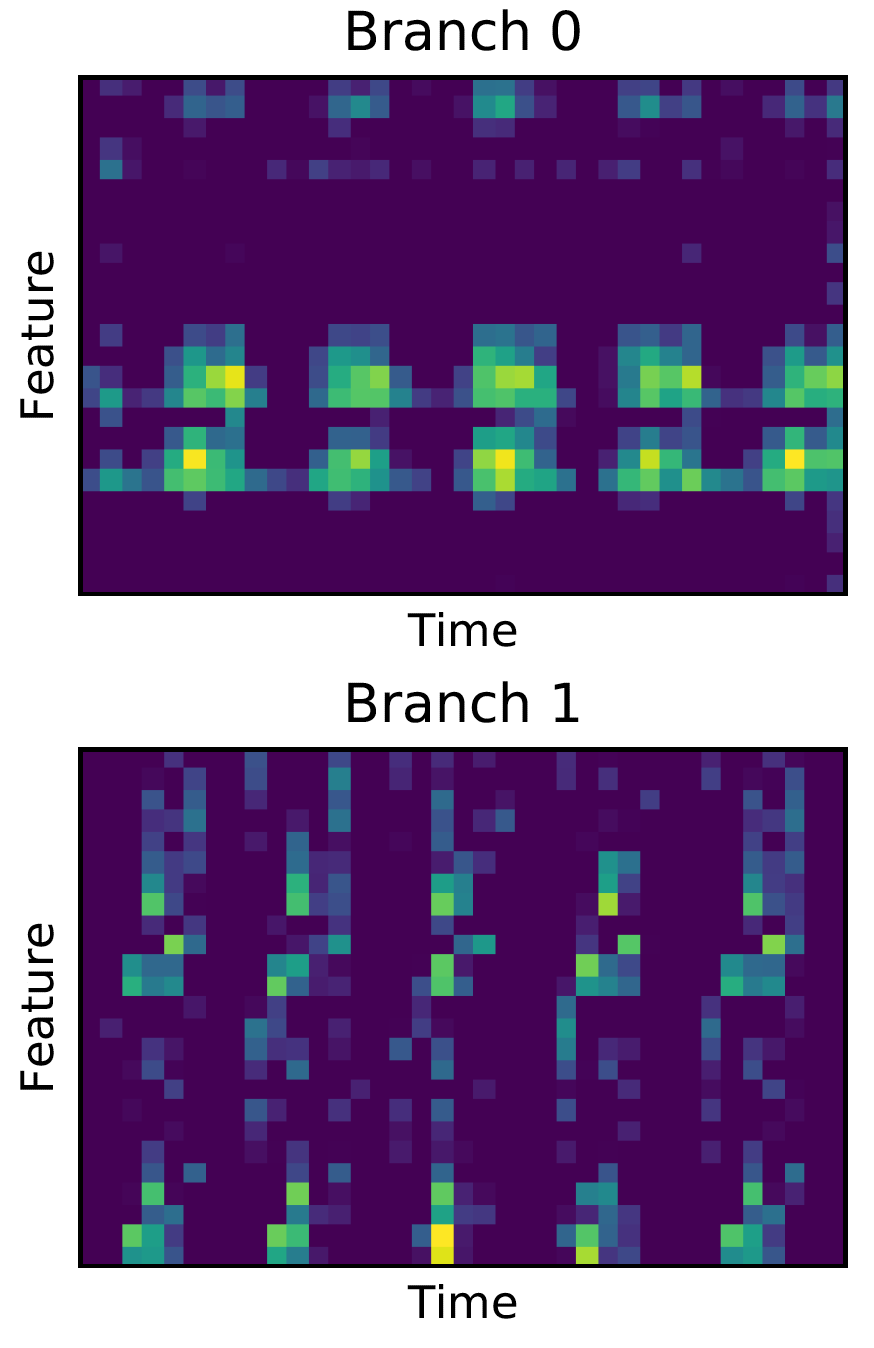}
  }
  \vspace{-3mm}
  \subfigure[SI output feature]
  {
      \includegraphics[width=0.45\linewidth]{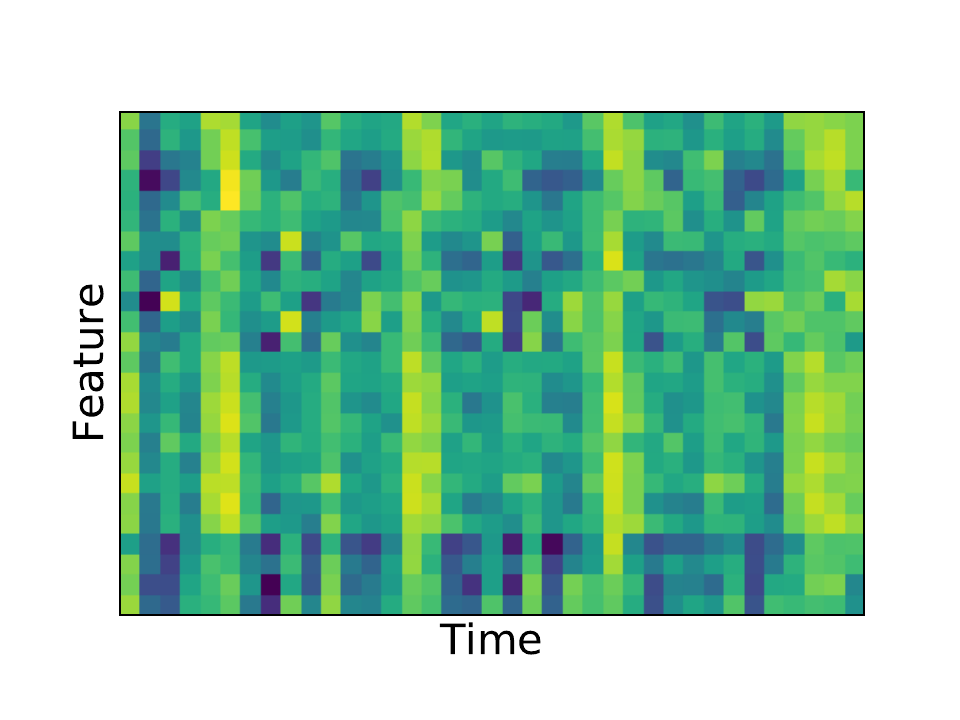}
  }
  \subfigure[TI output feature]
  {
      \includegraphics[width=0.45\linewidth]{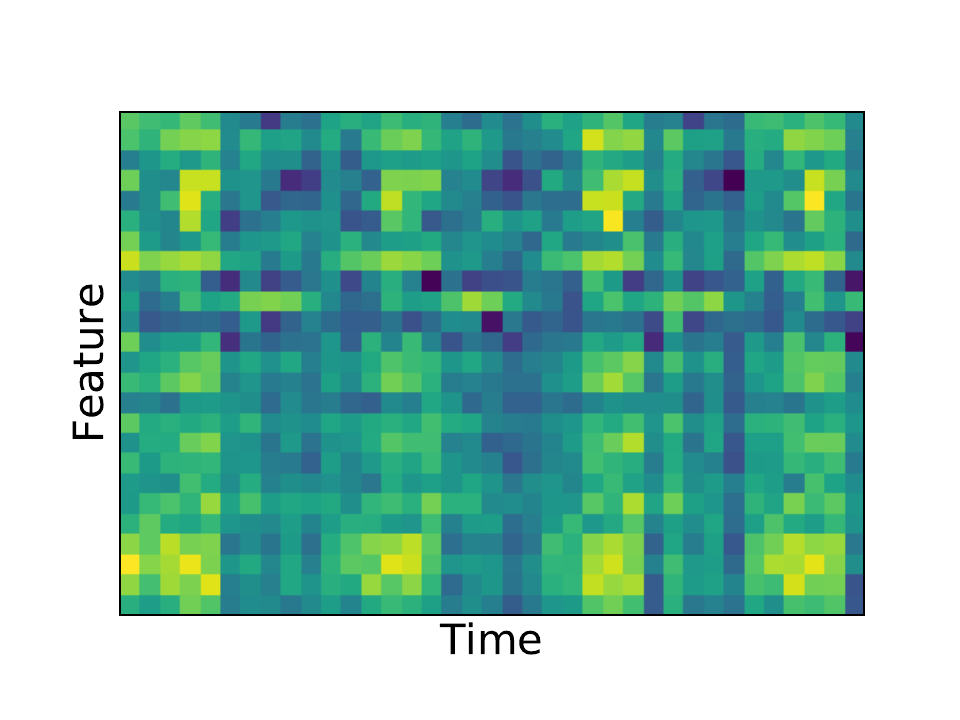}
  }
  \caption{The weight values of one random filter in the fusion
      modules of SI and TI paths, as well as the corresponding input/output features in the feature fusion module.
      (a) and (b) are the values of the 256-dimensional weights.  
      The horizontal axis
      denotes the input channels and each color denotes one branch.
      (c)-(f) are input and output feature maps with the maximum
      coefficient of different branches in SI and TI.
      The horizontal axis denotes the temporal dimension and the vertical axis denotes the feature channel dimension. 
      These figures show that STIGCN merges multiple order information and generates
      more informative representations.}
  \label{fig:vis_weight}
  \vspace{-3mm}
\end{figure}

One key insight of our spatio-temporal inception design is the fusion of
features from different branches, and each branch processes specific-order information.
To inspect how the feature fusion module synthesizes features of different
branches, we visualize the input feature maps, output feature maps and the
weights of 1$\times$1 convolutional layer in the fusion module of
SI and TI paths.
For convenience, we visualize the weights of one randomly selected filter, as well
as its input and output feature maps with the maximum coefficient, as shown in
Fig. \ref{fig:vis_weight}. In this figure, (a) and (b) demonstrate that every branch
contributes to the output; (c)-(f) indicate that different
branches learn discriminative features. Thus, we can obtain more informative outputs by synthesizing these features.
In general, STIGCN merges multiple sets of transformed information and benefits greatly from the aggregated representations.

\begin{figure*}[!ht]
  \centering
  \includegraphics[width=0.80\linewidth]{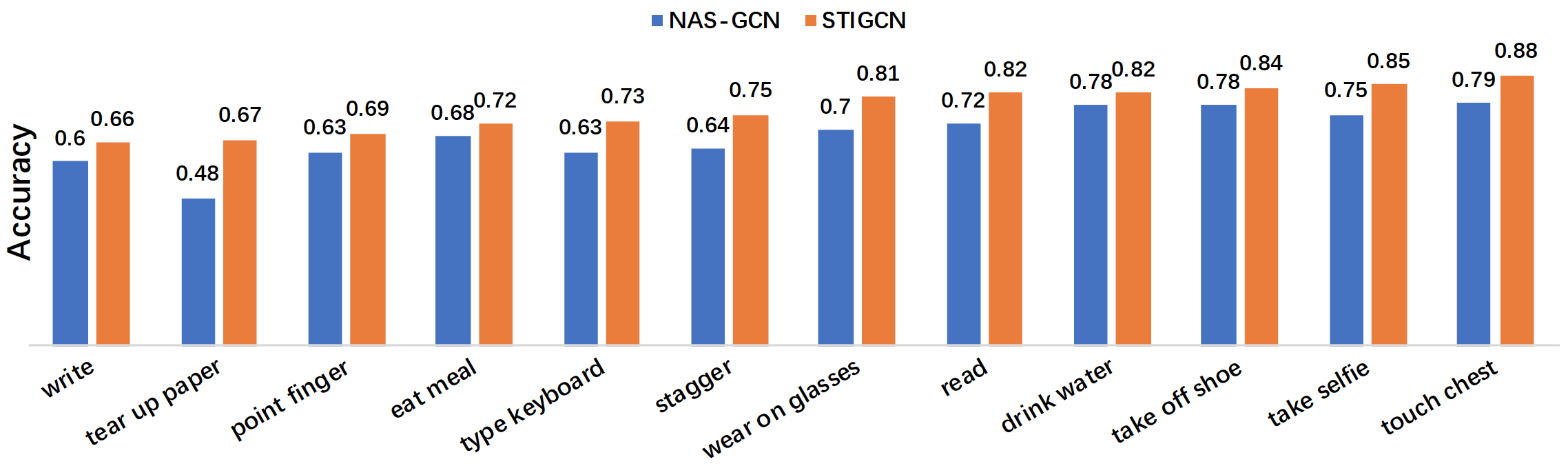}
  \vspace{-3mm}
  \caption{
  Comparison of classification accuracy of 12 difficult action classes on NTU RGB+D X-Sub.
  }
  \label{fig:bar}
\end{figure*}

\subsubsection{The Effect of the Representation Dimension.}
\begin{figure}[!t]
  \centering
  \includegraphics[width=0.65\linewidth]{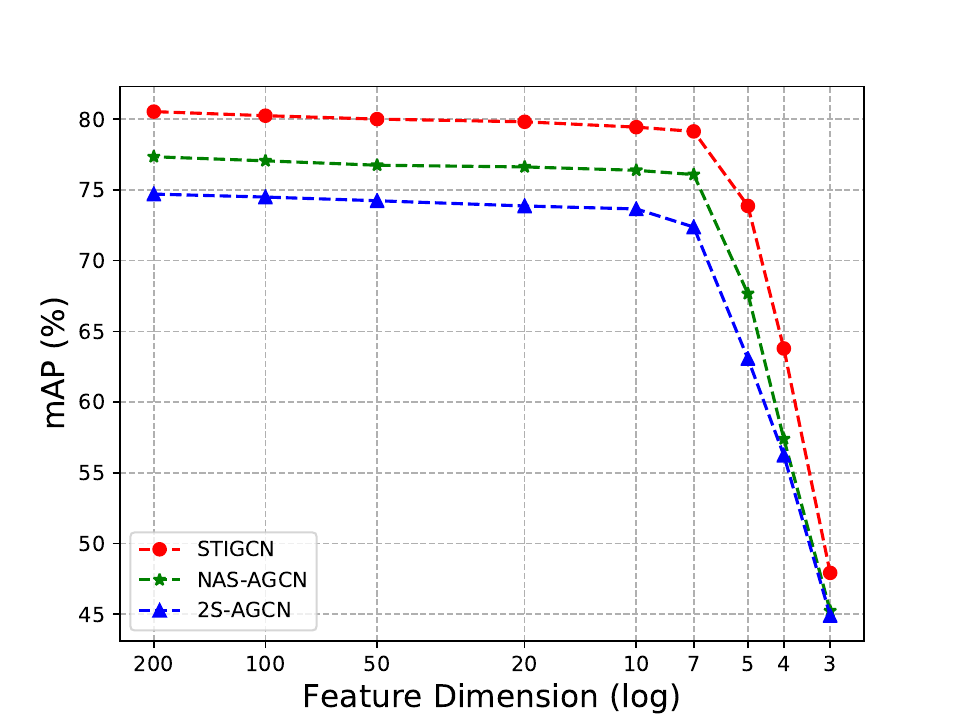}
  \vspace{-3mm}
  \caption{The evaluation of representations learned by
      different architectures with different dimensions.
  }
  \label{fig:pca}
  \vspace{-7mm}
\end{figure}

We reduce the dimension of output features of STIGCN, NAS-GCN and 2S-AGCN via
principal component analysis (PCA) and test their performance on the NTU RGB+D
retrieval task.
The results are shown in Fig. \ref{fig:pca}. The representation learned by
STIGCN consistently outperforms the others at varying dimensions from 200 to 3. An
interesting observation is that the performance of 2S-GCN and NAS-GCN tends to
be similar when the feature dimension is very small. STIGCN, in contrast,
outperforms these two models by a wide margin. This finding shows that with the
synthesis of multiple sets of transformed information, the representations from
STIGCN are more robust to the change of feature dimension.

\subsubsection{Embedding Representations.}
\begin{figure}[!t]
  \centering
  \subfigure[NAS-GCN]
  {
      \includegraphics[width=0.47\linewidth]{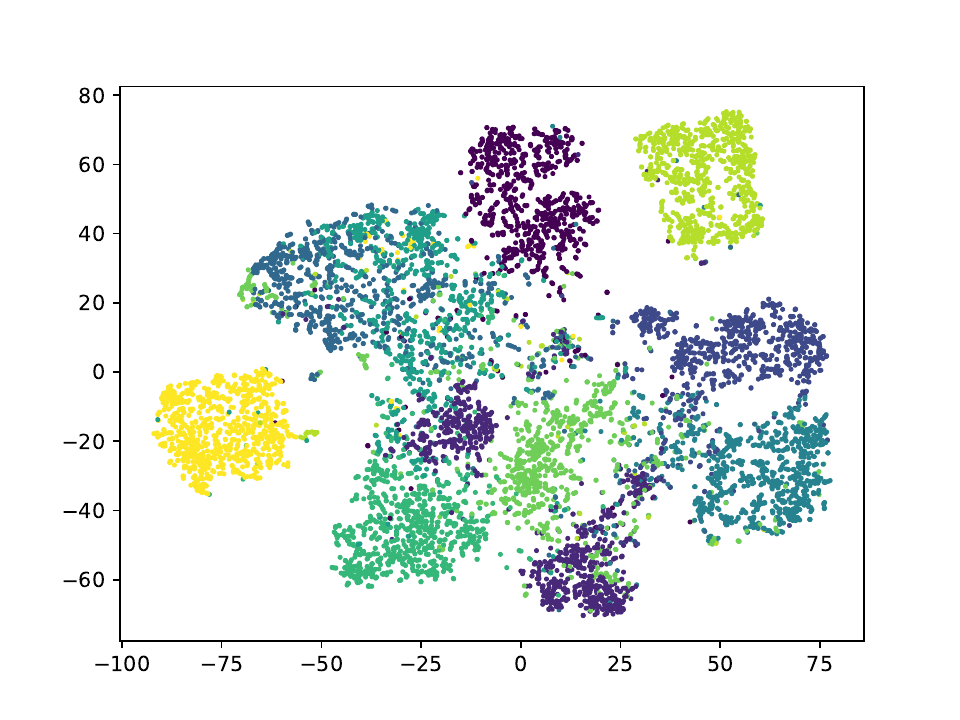}
  }
  \subfigure[STIGCN]
  {
      \includegraphics[width=0.47\linewidth]{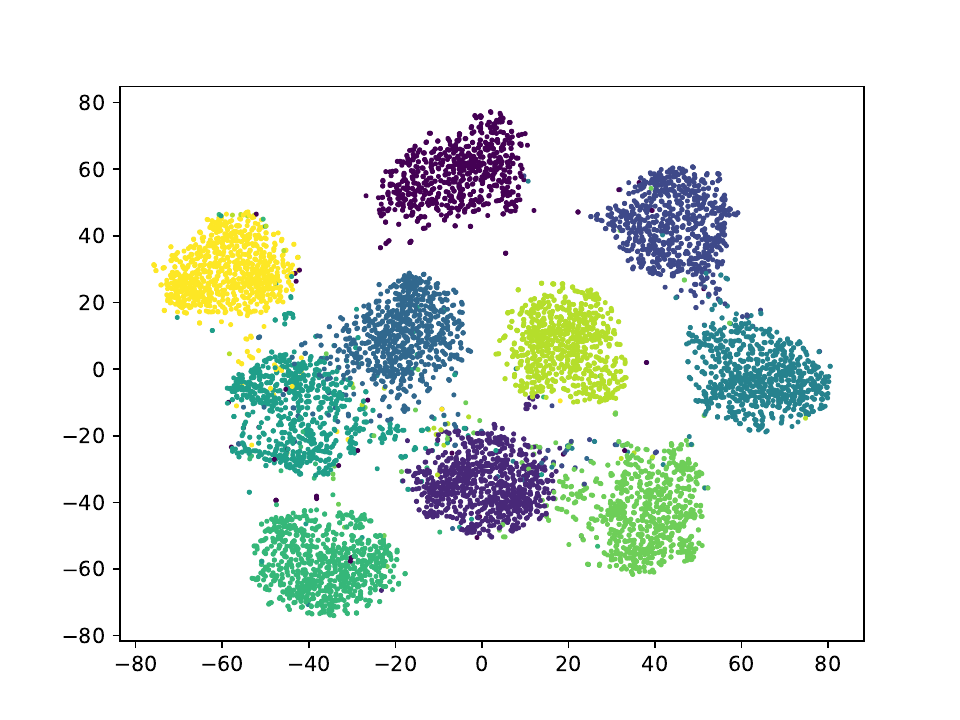}
  }
  \vspace{-5mm}
  \caption{
  Visualizations of skeleton-sequence representation embeddings.
  Each sequence is visualized as one point, and the colors denote different action classes. 
  }
  \label{fig:tsne}
  \vspace{-5mm}
\end{figure}

Fig. \ref{fig:tsne} further shows the t-SNE \cite{tsne} visualization of
the embedding of skeleton-sequence representations learned from NAS-GCN and STIGCN. We
use the testing set of NTU RGB+D and the output representations are
projected into a 2-dimensional space using t-SNE.
This figure clearly shows that representations generated by STIGCN are semantically better grouped than those of
NAS-GCN.

\subsubsection{Classification Accuracy on Difficult Actions}

We further analyze the performance of NAS-GCN and STIGCN on difficult action classes, i.e., actions whose classification accuracy is less than $80\%$ in either NAS-GCN or STIGCN.
As shown in Fig. \ref{fig:bar}, there are 12 difficult actions for NAS-GCN and 6 difficult actions for STIGCN. 
STIGCN outperforms 2S-AGCN on all these difficult actions. This finding shows that STIGCN has better ability to handle challenging actions.

\section{Conclusion}
\label{sec:conclusions}
In this paper, we propose a simple graph convolution backbone architecture
called spatial temporal inception graph convolutional networks for skeleton-based action recognition.
It overcomes the limitations of previous methods in extracting and
synthesizing information of different scales and transformations from
different paths at different levels.
On two large-scale datasets, the
proposed network outperforms state-of-the-art methods by a significant margin with
surprisingly fewer parameters and FLOPs.
Our method indicates that increasing the number of sets of transformations is a more effective way of gaining accuracy than simply
creating wider GCNs. We hope this insight will facilitate the iteration of
GCN-based backbones for spatio-temporal sequence analyses.
In the future, we will explore more types of transformations for the design of
graph convolution building blocks.

\begin{acks}
This work was partially supported by Major Scientific Research Project of
Zhejiang Lab (No. 2019DB0ZX01), National Key Research and Development
Program of China under Grant 2017YFB1002203 and the National Natural Science Foundation of
China under Grant 61872329.
\end{acks}

\bibliographystyle{ACM-Reference-Format}
\bibliography{draft}


\end{document}